\documentclass[10pt,twocolumn,letterpaper]{article}

\usepackage{cvpr}
\usepackage{times}
\usepackage{epsfig}
\usepackage{graphicx}
\usepackage{amsmath}
\usepackage{amssymb}
\usepackage{bm}
\usepackage{graphicx}
\usepackage{subfig}
\usepackage{float}  %
\usepackage{tabularx}
\usepackage[export]{adjustbox}
\usepackage{amsmath} %
\usepackage{animate} %
\usepackage{float} %

\usepackage[pagebackref=true,breaklinks=true,letterpaper=true,colorlinks,bookmarks=false]{hyperref}

\cvprfinalcopy %

\def\cvprPaperID{****} %

\ifcvprfinal\fi
\begin{document}

\title{Visual Data Augmentation through Learning}

\author{
Grigorios G. Chrysos$^{1}$, \quad Yannis Panagakis$^{1,2}$, \quad Stefanos Zafeiriou$^{1}$ \\ 
{\textsuperscript{1} Department of Computing, Imperial College London, UK}\\
{\textsuperscript{2} Department of Computer Science, Middlesex University London, UK}\\
{\texttt{\{g.chrysos, i.panagakis, s.zafeiriou\}@imperial.ac.uk}}
}

\maketitle

\begin{abstract}
  The rapid progress in machine learning methods has been empowered by i) huge datasets that have been collected and annotated, ii) improved engineering (e.g. data pre-processing/normalization). The existing datasets typically include several million samples, which constitutes their extension a colossal task. In addition, the state-of-the-art data-driven methods demand a vast amount of data, hence a standard engineering trick employed is artificial data augmentation
  for instance by adding into the data cropped and  (affinely) transformed images. However, this approach does not correspond to any change in the natural 3D scene.
  We propose instead to perform data augmentation through learning realistic local transformations. We learn a forward and an inverse transformation that maps an image from the high-dimensional space of pixel intensities to a latent space
  which varies (approximately) linearly with the latent space of a realistically transformed version of the image. Such transformed images can be considered two successive frames in a video. Next,
   we utilize these transformations to learn a linear model that modifies the latent spaces and then use the inverse transformation to synthesize a new image. We argue that the this procedure produces powerful invariant representations. We perform both qualitative and quantitative experiments that demonstrate our proposed method creates new realistic images.
\end{abstract}

\section{Introduction}
\label{sec:linear_dynamics_introduction}

\begin{figure}[!t]
    \centering
    \includegraphics[width=1\linewidth]{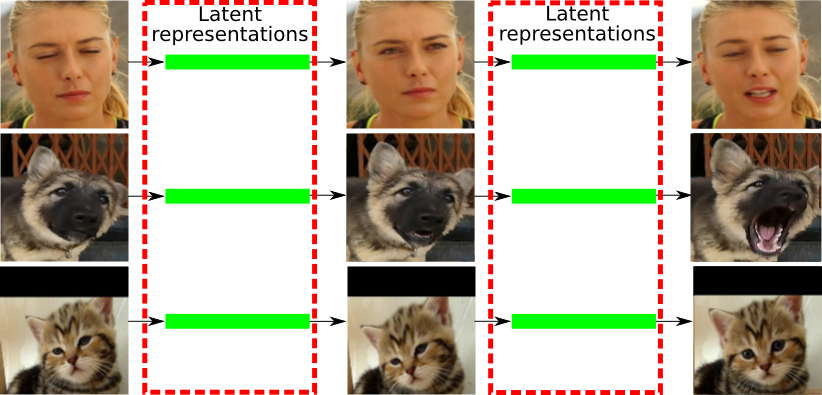}
    \caption{(Preferably viewed in color) We want to augment arbitrary images (left column) by learning local transformations. We find a low-dimensional space and learn the forward and inverse transformations from the image to the representation space. Then, we can perform a simple linear transformation in the (approximately) linear low-dimensional space and acquire a new synthesized image (the middle column). The same procedure can be repeated with the latest synthesized image (e.g. from the middle to the right columns).}
    \label{fig:linear_dynamics_motivational_fig}
\end{figure}

The lack of training data has till recently been an impediment for training machine learning methods. The latest breakthroughs of Neural Networks (NNs) can be partly attributed to the increased amount of (public) databases with massive number of labels/meta-data. Nevertheless, the state-of-the-art networks include tens or hundreds of millions of parameters~\cite{he2015deep, zoph2017learning}, i.e. they require more labelled examples than we have available. %
To ameliorate the lack of sufficient labelled examples, different data augmentation methods have become commonplace during training. 
In this work, we propose a new data augmentation technique that finds a low-dimensional space in which performing a simple linear change results in a nonlinear change in the image space.

Data augmentation methods are used as label-preserving transformations with twofold goals: a) avoid over-fitting, b) ensure that enough samples have been provided to the network for learning. 
A plethora of label-preserving transformations have been proposed, however the majority is classified into a) either model-based methods, b) or generic augmentations. The model-based demand an elaborate model to augment the data, e.g. the 3DMM-based~\cite{blanz1999morphable} face profiling of \cite{tran2017disentangled}, the novel-view synthesis from 3D models~\cite{rematas2014image}. Such models are available for only a small number of classes and the realistic generation from 3D models/synthetic data is still an open problem~\cite{bousmalis2016unsupervised}. The second augmentation category is comprised of methods defined artificially; these methods do not correspond to any natural movement in the scene/object. For instance, a 2D image rotation does not correspond to any actual change in the 3D scene space; it is purely a computational method for encouraging rotation invariance.

We argue that a third category of augmentations consists of local transformations. We learn a nonlinear transformation that maps the image to a low-dimensional space that is assumed to be (approximately) linear. This linear property allows us to perform a linear transformation and map the original latent representation to the representation of a slightly transformed image (e.g. a pair of successive frames in a video). If we can learn the inverse transform, i.e. mapping from the low-dimensional space to the transformed image, then we can modify the latent representation of the image linearly and this results in a nonlinear change in the image domain. 

We propose a three-stage approach that learns a forward transformation (from image to low-dimensional representation) and an inverse transformation (from latent to image representation) so that a linear change in the latent space results in a nonlinear change in the image space.
The forward and the inverse learned transformations are approximated by an Adversarial Autoencoder and a GAN respectively. 

\begin{figure}[!hbp]
    \centering
    \includegraphics[width=0.495\linewidth]{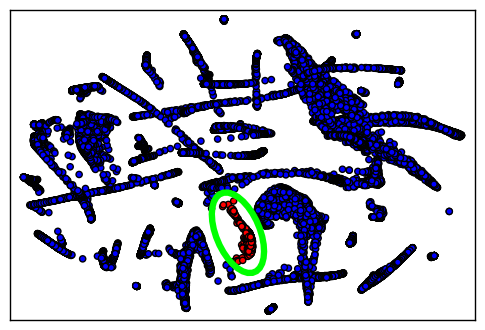}
    \includegraphics[width=0.495\linewidth]{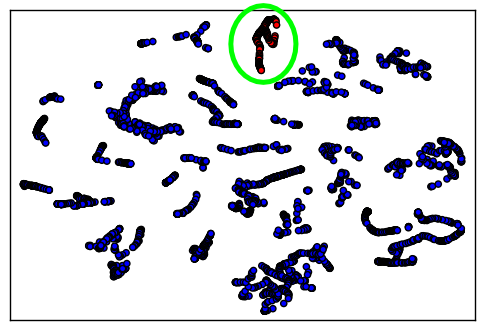}
    \caption{(Preferably viewed in color) T-SNE~\cite{maaten2008visualizing} in the latent representations of a) 300VW~\cite{shen2015first} (left Fig.), b) cats' videos. In both cases the representations of the first video (red dots) are compared against the rest videos (blue dots)). To avoid cluttering the graphs every second frame is skipped (their representation is similar to the previous/next frame). For further emphasis, a green circle is drawn around the red points.}
    \label{fig:linear_dynamics_tsne}
\end{figure}

In our work, we learn object-specific transformations while we do not introduce any temporal smoothness. Even though learning a generic model for all classes is theoretically plausible, we advocate that with the existing methods, there is not sufficient capacity to learn such generic transformations for all the objects. Instead we introduce object-specific transformations.  
Even though we have not explicitly constrained our low-dimensional space to be temporally smooth, e.g. by using the cosine distance, we have observed that the transformations learned are powerful enough to linearize the space. As a visual illustration, we have run T-SNE~\cite{maaten2008visualizing} with the  latent representations of the first video of 300VW~\cite{shen2015first} against the rest 49 videos of the published training set; Fig.~\ref{fig:linear_dynamics_tsne} validates our hypothesis that the latent representations of that video reside in a discrete cluster over the rest of the representations. In a similar experiment with the collected videos of cats, the same conclusion is reached, i.e. the representations of the first video form a discrete cluster.

We have opted to report the results in the facial space that is highly nonlinear, while the representations are quite rich. 
To assess further our approach, we have used two ad-hoc objects, i.e. cat faces, dog faces, that have far less data labelled available online. Additionally, in both ad-hoc objects the shape/appearance presents greater variation than that of human faces, hence more elaborate transformations should be learned.

In the following Sections we review the neural networks based on which we have developed our method (Sec.~\ref{sec:linear_dynamics_describe_networks}), introduce our method in Sec.~\ref{sec:linear_dynamics_method}. Sequentially, we demonstrate our experimental results in Sec.~\ref{sec:linear_dynamics_experiment}. Due to the restricted space, additional visualizations are deferred to the supplementary material, including indicative figures of the cats', dogs' videos, additional (animated) visual results of our method, an experiment illustrating that few images suffice to learn object-specific deformable models. We strongly encourage the reviewers to check the supplementary material.

\textbf{Notation:} A small (capital) bold letter represents a vector (matrix); a plain letter designates a scalar number. A vectorized image of a dynamic scene at time $t$ is denoted as $\bm{i}^{(t)}$, while $\bm{i}_k^{(t_k)}$ refers to the $k^{th}$ training sample.

\section{Background}
The following lines of research are related with our proposed method:

\label{sec:linear_dynamics_related_work}
\textbf{Model-based augmentation for faces}: The methods in this category utilize 2D/3D geometric information. In T-CNN~\cite{wu2015facial} the authors introduce an alignment-sensitive method tailored to their task. Namely, they warp a face from its original shape (2D landmarks) to a similar shape (based on their devised clustering).  Recently, Zhu \etal~\cite{zhu2016face} use a 3D morphable model (3DMM)~\cite{blanz1999morphable} to simulate the effect of profiling for synthesizing images in large poses. Tran \etal in \cite{tran2017disentangled} fit a 3DMM to estimate the facial pose and learn a GAN conditioned on the pose. During inference, new facial images are synthesized by sampling different poses. The major limitation of the model-based methods is that they require elaborate 2D/3D models. Such models have been studied only for the human face\footnote{18 years since the original 3DMM model and the problem is not solved for all cases.} or the human body, while the rest objects, e.g. animals faces, have not attracted such attention yet. On the contrary, our method is not limited to any object (we have learned models with cats' faces and dogs' faces) and does not require elaborate 3D/2D shape models.

\textbf{Unconditional image synthesis}: The successful application of GANs~\cite{goodfellow2014generative} in a variety of tasks including photo-realistic image synthesis\cite{ledig2016photo}, style transfer~\cite{yoo2016pixel}, inpainting~\cite{pathak2016context}, image-to-image mapping tasks~\cite{isola2016image} has led to a proliferation of works on unconditional image synthesis~\cite{arjovsky2017wasserstein, zhao2016energy}. Even though unconditional image generation has significant applications, it cannot be used for conditional generation when labels are available. Another line of research is directly approximating the conditional distribution over pixels~\cite{oord2016pixel}. The generation of a single pixel is conditioned on all the previously generated pixels. Even though realistic samples are produced, it is costly to sample from them; additionally such models do not provide access to the latent representation.

\textbf{Video frames' prediction}: The recent (experimental) breakthroughs of generative models have accelerated the progress in video frames prediction. In \cite{vondrick2016generating} the authors learn a model that captures the scene dynamics and synthesizes new frames. To generalize the deterministic prediction of \cite{vondrick2016generating}, the authors of \cite{xue2016visual} propose a probabilistic model, however they show only a single frame prediction in low-resolution objects. In addition, the unified latent code $\bm{z}$ (learned for all objects) does not allow particular motion patterns, e.g. of an object of interest in the video, to be distinguished. Lotter \etal~\cite{lotter2016deep} approach the task as a conditional generation. They employ a Recurrent Neural Network (RNN) to condition future frames on previously seen ones, which implicitly imposes temporal smoothness. 
A core differentiating factor of these approaches from our work is that they i) impose temporal smoothness, ii) make simplifying assumptions (e.g. stationary camera~\cite{vondrick2016generating}); these restrictions constrain their solution space and allow for realistic video frames' prediction. In addition, the techniques for future prediction often result in blurry frames, which can be attributed to the multimodal distributions of unconstrained natural images, however our end-goal consists in creating realistic images for highly-complex images, e.g. animals' faces.

The work of \cite{goroshin2015learning} is the most similar to our work. The authors construct a customized architecture and loss to linearize the feature space and then perform frame prediction to demonstrate that they have successfully achieved the linearization. Their highly customized architecture (in comparison to our off-the-shelves networks) have not been applied to any highly nonlinear space, in \cite{goroshin2015learning} mostly synthetic, simple examples are demonstrated. Apart from the highly nonlinear objects we experiment with, we provide several experimental indicators that our proposed method achieves this linearization in challenging cases.

An additional differentiating factor from the aforementioned works is that, to the best of our knowledge, this three-stage approach has not been used in the past for a related task.

\subsection{cGAN and Adversarial Autoencoder}
\label{sec:linear_dynamics_describe_networks}
Let us briefly describe the two methods that consist our workhorse for learning the transformations. These are the conditional GAN and the Adversarial Autoencoder.

A Generative Adversarial Network (GAN)~\cite{goodfellow2014generative} is a generative network that has been very successfully employed for learning probability distributions~\cite{ledig2016photo}. A GAN is comprised of a generator $G$ and a discriminator $D$ network, where the generator samples from a pre-defined distribution in order to approximate the probability distribution of the training data, while the discriminator tries to distinguish between the samples originating from the model distribution to those from the data distribution. 
\textbf{Conditional GAN (cGAN)}~\cite{mirza2014conditional} extends the formulation by conditioning the distributions with additional labels. More formally, if we denote with $p_{d}$ the true distribution of the data, with $p_{\bm{z}}$ the distribution of the noise, with $\bm{s}$ the conditioning label and $\bm{y}$ the data, then the objective function is:

\begin{equation}
\begin{split} %
    \mathcal{L}_{cGAN}(G, D) = \mathbb{E}_{\bm{s},\bm{y} \sim p_{d}(\bm{s},\bm{y})}[\log D(\bm{s},\bm{y})] + \\
    \mathbb{E}_{\bm{s} \sim p_{d}(\bm{s}), \bm{z} \sim p_{z}(\bm{z})}[\log (1-D(\bm{s},G(\bm{s},\bm{z})))]
    \label{eq:linear_dynamics_cgan_loss}
\end{split}
\end{equation}

This objective function is optimized in an iterative manner, as 
\begin{equation}
\begin{split} %
    \min_{\bm{w}_G} \max_{\bm{w}_D} \mathcal{L}_{cGAN}(G, D) = \mathbb{E}_{\bm{s},\bm{y} \sim p_{d}(\bm{s},\bm{y})}[\log D(\bm{s},\bm{y}; \bm{w}_D)] + \\
    \mathbb{E}_{\bm{s} \sim p_{d}(\bm{s}), \bm{z} \sim p_{z}(\bm{z})}[\log (1-D(\bm{s},G(\bm{s},\bm{z}; \bm{w}_G); \bm{w}_D))]
    \nonumber
\end{split}
\end{equation}

where $\bm{w}_G, \bm{w}_D$ denote the generator's, discriminator's parameters respectively.

An Autoencoder (AE)~\cite{hinton1994autoencoders, masci2011stacked} is a neural network with two parts (an encoder and a decoder) and aims to learn a latent representation $\bm{z}$ of their input $\bm{y}$. Autoencoders are mostly used in an unsupervised learning context~\cite{kan2014stacked} with the loss being the reconstruction error. On the other hand, an \textbf{Adversarial Autoencoder (AAE)}~\cite{makhzani2015adversarial} consists of two sub-networks: i) a generator (an AE network), ii) a discriminator. The discriminator, which is motivated by GAN's discriminator, accepts the latent vector (generated by the encoder) and tries to match the latent space representation with a pre-defined distribution.

\section{Method}
\label{sec:linear_dynamics_method}
\begin{figure*}[!h]
    \centering
    \includegraphics[width=1\linewidth]{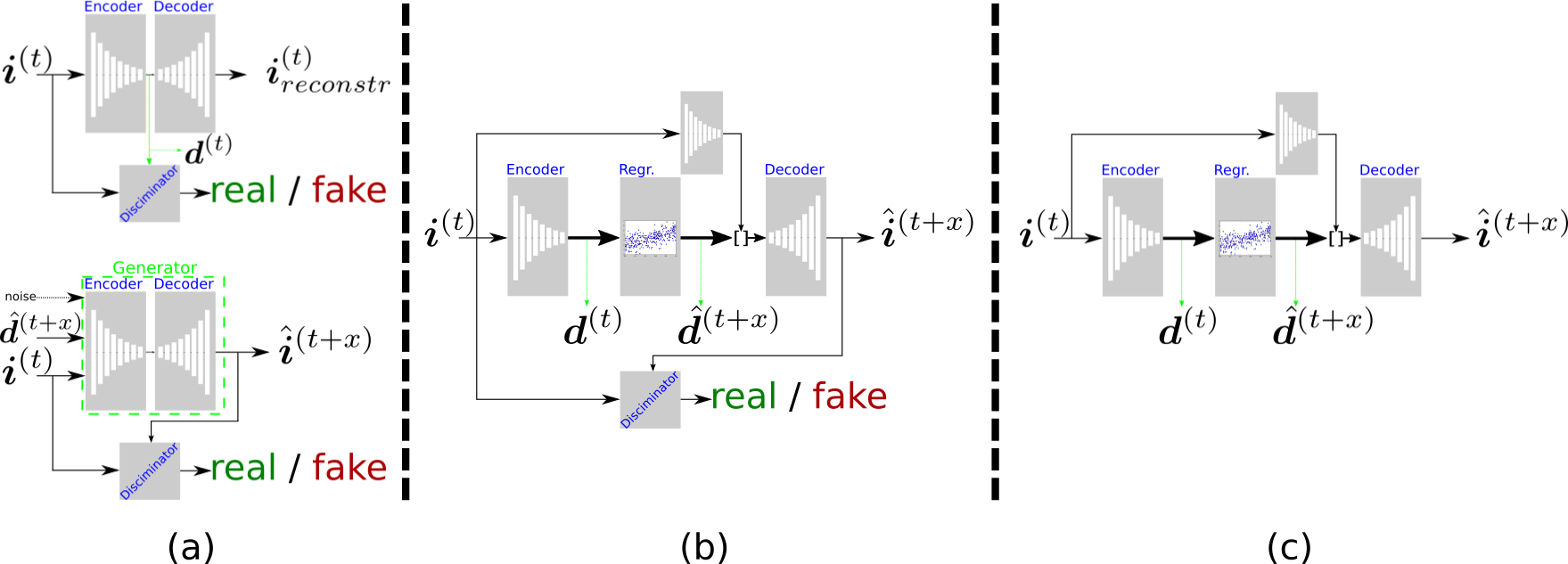}
    \caption{The architectures used in (a) separate training per step (the network for Stage I is on the top, for Stage III on the bottom), (b) fine-tuning of the unified model, (c) prediction. The `[]' symbol denotes concatenation.}
    \label{fig:linear_dynamics_network_structure} %
\end{figure*}

The core idea of our approach consists in finding a low-dimensional space that is (approximately)  linear with respect to the projected representations. We aim to learn the (forward and inverse) transformations from the image space to the low-dimensional space. We know that an image $\bm{i}^{(t)}$ is an instance of a dynamic scene at time t, hence the difference between the representations of two temporally close moments should be small and linear. We can learn the linear transitions of the representations and transform our image to $\bm{i}^{(t + x)}$. We perform this linearization in 2-steps; an additional step is used to synthesize images of the same object with slightly different representations. The synthesized image can be thought of as a locally transformed image, e.g. the scene at $t + x$ moment with $x$ sufficiently small.

\subsection{Stage I: Latent image representation}
\label{ssec:linear_dynamics_stageI}
Our goal consists in learning the transformations to the linearized space, however for the majority of the objects there are not enough videos annotated that can express a sufficient percent of the variation. For instance, it is not straightforward to find long videos of all breeds of dogs where the full body is visible. However, there are far more static images available online, which are faster to collect and can be used to learn the transformation from the image space to the latent space.

In an unsupervised setting a single image $\bm{i}^{(t)}$ (per step) suffices for learning latent representations, no additional labels are required, which is precisely the task that Autoencoders were designed for. The latent vector of the Autoencoder lies in the latent space we want to find.

We experimentally noticed that the optimization converged faster if we used an adversarial learning procedure. We chose an Adversarial Autoencoder (AAE)~\cite{makhzani2015adversarial} with a customized loss function. The encoder $f_e^{I}$ accepts an image $\bm{i}^{(t)}$, encodes it to $\bm{d}^{(t)}$; the decoder $f_d^{I}$ reconstructs $\bm{i}^{(t)}$. We modify the discriminator to accept both the latent representation and the reconstructed image as input (fake example) and try to distinguish those from the distribution sample and the input image respectively. Moreover, we add a loss term that captures the reconstruction loss, which in our case consists of i) an $\ell_1$ norm and ii) $\ell_1$ in the image gradients.
Consequently, the final loss function is comprised of the following two terms: i) the adversarial loss, ii) the reconstruction loss or:
\begin{equation}
    \mathcal{L}^{I} = \mathcal{L}_{adver} + \lambda_{I} \mathcal{L}_{rec}^{I}
\end{equation}
with
\begin{equation}
    \mathcal{L}_{rec}^{I} = ||f_d^{I}(f_e^{I}(\bm{y})) - \bm{y}||_{\ell_1} + ||\nabla f_d^{I}(f_e^{I}(\bm{y})) - \nabla \bm{y}||_{\ell_1}
    \label{eq:linear_dynamics_stageI_recloss}
\end{equation}

The vector $\bm{y}$ in this case is a training sample $\bm{i}_k^{(t_k)}$, while $\lambda_{I}$ is a hyper-parameter.

\subsection{Stage II: Linear Model Learning}
\label{ssec:linear_dynamics_stageII}
In this stage the latent representation $\bm{d}^{(t)}$ of an image $\bm{i}^{(t)}$ (as learned from stage I) is used to learn a mapping to the latent representation $\bm{d}^{(t + x)}$ of the image $\bm{i}^{(t + x)}$; the simple method of linear regression is chosen as a very simple transformation we can perform in a linear space. Given N pairs of images\footnote{Each pair includes two highly correlated images, i.e. two nearby frames from a video sequence.} $\{(\bm{i}_1^{(t_1)}, \bm{i}_1^{(t_1 + x_1)}), (\bm{i}_2^{(t_2)}, \bm{i}_2^{(t_2 + x_2)}), \ldots, (\bm{i}_N^{(t_N)}, \bm{i}_N^{(t_N + x_N)})\}$, the set of the respective latent representations $\bm{\mathcal{D}} = \{(\bm{d}_1^{(t_1)}, \bm{d}_1^{(t_1 + x_1)}), (\bm{d}_2^{(t_2)}, \bm{d}_2^{(t_2 + x_2)}), \ldots, (\bm{d}_N^{(t_N)}, \bm{d}_N^{(t_N + x_N)})\}$; the set $\bm{\mathcal{D}}$ is used to learn the linear mapping:
\begin{equation}
    \bm{d}^{(t_j + x_j)} = \bm{A}\cdot[\bm{d}^{(t_j)}; 1] + \bm{\epsilon}
\end{equation}

where $\bm{\epsilon}$ is the noise; the Frobenius norm of the residual consists the error term: 
\begin{equation}
    L = ||\bm{d}^{(t_j + x_j)} - \bm{A}\cdot[\bm{d}^{(t_j)}; 1]||_F^2
    \label{eq:linear_dynamics_lin_regr_basic}
\end{equation}

To ensure the stability of the linear transformation we add a Tikhonov regularization term (i.e, Frobenius norm)  on Eq.~\ref{eq:linear_dynamics_lin_regr_basic}. That is,
\begin{equation}
    \mathcal{L}^{II} = ||\bm{d}^{(t_j + x_j)} - \bm{A}\cdot[\bm{d}^{(t_j)}; 1]||_F^2 + \lambda_{II} ||\bm{A}||_f^2,
    \label{eq:linear_dynamics_lin_regr_regul}
\end{equation}
with $\lambda_{II}$ a regularization hyper-parameter. The closed-form solution to Eq.~\ref{eq:linear_dynamics_lin_regr_regul} is 
\begin{equation}
    \bm{A} = \bm{Y} \cdot \bm{X}^T \cdot (\bm{X}\cdot\bm{X}^T + \lambda_{II} \cdot \bm{\mathcal{I}})^{-1},
    \label{eq:linear_dynamics_lin_regr_solution_a}
\end{equation}
where $\bm{\mathcal{I}}$ denotes an identity matrix, $\bm{X}$, $\bm{Y}$ two matrices that contain column-wise the initial and target representations respectively, i.e. for the $k^{th}$ sample  $\bm{X}(:, k) = [\bm{d}_k^{(t_k)}; 1]$, $\bm{Y}(:, k) = \bm{d}_k^{(t_k + x_k)}$.

\subsection{Stage III: Latent representation to image}
\label{ssec:linear_dynamics_stageIII}
In this step, we want to learn a transformation from the latent space to the image space, i.e. the inverse transformation of Stage I. In particular, we aim to map the regressed representation $\hat{\bm{d}}^{(t + x)}$ to the image $\bm{i}^{(t + x)}$. Our prior distribution consists of a low-dimensional space, which we want to map to a high-dimensional space; GANs have experimentally proven very effective in such mappings~\cite{ledig2016photo, pathak2016context}. 

A conditional GAN is employed for this step; we condition GAN in both the (regressed) latent representation $\hat{\bm{d}}^{(t + x)}$ and the original image $\bm{i}^{(t)}$. Conditioning on the original image has experimentally resulted in faster convergence and it might be a significant feature in case of limited amount of training samples. Inspired by the work of \cite{isola2016image}, we form the generator as an autoencoder denoting the encoder as $f_e^{III}$, the decoder as $f_d^{III}$. Skip connections are added from the second and fourth layers of the encoder to the respective layers in the decoder with the purpose of allowing the low-level features of the original images to be propagated to the result. 

In conjunction with \cite{isola2016image} and Sec.~\ref{ssec:linear_dynamics_stageI}, we add a reconstruction loss term as 
\begin{equation}
    \mathcal{L}_{rec}^{III} = ||f_d^{III}(f_e^{III}(\bm{y})) - \bm{s}||_{\ell_1} + ||\nabla f_d^{III}(f_e^{III}(\bm{y})) - \nabla \bm{s}||_{\ell_1}
    \label{eq:linear_dynamics_stageIII_recloss}
\end{equation}
where $\bm{y}$ is a training sample $\bm{i}_k^{(t_k)}$ and $\bm{s}$ is the conditioning label (original image) $\bm{i}_{k - x}^{(t_{k - x})}$.
In addition, we add a loss term that encourages the features of the real/fake samples to be similar. Those features are extracted from the penultimate layer of the AAE's discriminator. Effectively, this leads the fake (i.e. synthesized) images to have representations that are close to the original image. 
The final objective function for this step includes three terms, i.e. the adversarial, the reconstruction and the feature loss:

\begin{equation}
    \mathcal{L}^{III} = \mathcal{L}_{cGAN} + \lambda_{III} \mathcal{L}_{rec}^{III} + \lambda_{III, feat} \mathcal{L}_{feat}
\end{equation}

\begin{figure*}[!t]
    \centering
    \subfloat[][Human faces] {
        \includegraphics[width=0.325\linewidth]{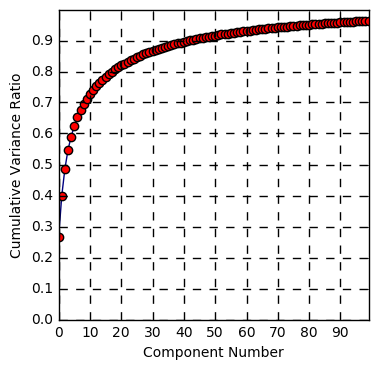}}
    \subfloat[][Cats' faces] {
    \includegraphics[width=0.325\linewidth]{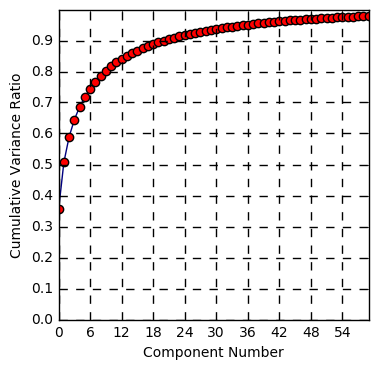}}
    \subfloat[][Dogs' faces] {
    \includegraphics[width=0.325\linewidth]{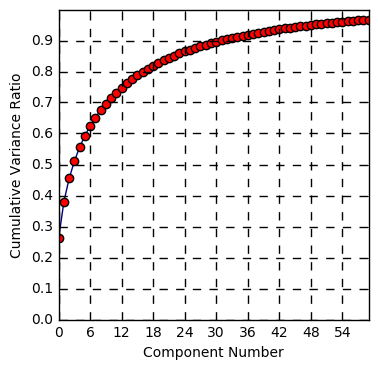}}
    \caption{Average variance in the dynamics representation per video for (a) the case of human faces, (b) cats' faces, (c) dogs' faces.}
    \label{fig:linear_dynamics_pca_variance}
\end{figure*}
\begin{figure*}[!t]
    \centering
    \captionsetup[subfigure]{labelformat=empty} %
     \subfloat[][] {
    \animategraphics[loop,autoplay,width=0.16\linewidth]{1}{figures/linear_dynamics/visualization_for_sequential_calls/1049105_1942-03-23_2009_}{0}{5}
    \includegraphics[width=0.16\linewidth]{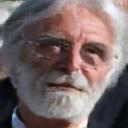}
    \includegraphics[width=0.16\linewidth]{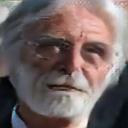}
    \includegraphics[width=0.16\linewidth]{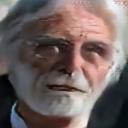}
    \includegraphics[width=0.16\linewidth]{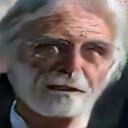}
    \includegraphics[width=0.16\linewidth]{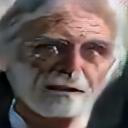}
    } \\[-2ex]
     \subfloat[][] {
    \animategraphics[loop,autoplay,width=0.16\linewidth]{1}{figures/linear_dynamics/visualization_for_sequential_calls/10043078_1954-03-18_2006_}{0}{5}
    \includegraphics[width=0.16\linewidth]{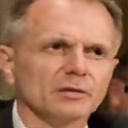}
    \includegraphics[width=0.16\linewidth]{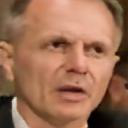}
    \includegraphics[width=0.16\linewidth]{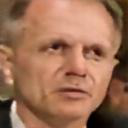}
    \includegraphics[width=0.16\linewidth]{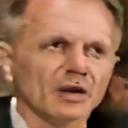}
    \includegraphics[width=0.16\linewidth]{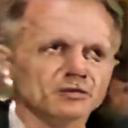}
    }
    \caption{(Preferably viewed in color) Conditional, iterative prediction from our proposed method. The images on the left are the original ones; then from the left to the right the $i^{th}$ column depicts the $(i - 1)^{th}$ synthesized image (iteration $(i - 1)$). In both rows, the image on the left is animated, hence if opened with Adobe Acrobat reader the transitions will be auto-played.}
    \label{fig:linear_dynamics_sequential_executions}
\end{figure*}

where $\mathcal{L}_{cGAN}$ is defined in Eq.\ref{eq:linear_dynamics_cgan_loss}, $\mathcal{L}_{feat}$ represents the similarity cost imposed on the features from the discriminator's penultimate layer and $\lambda_{III}$, $\lambda_{III, feat}$ are scalar hyper-parameters. To reduce the amount of hyper-parameters in our work, we have set $\lambda_{III} = \lambda_{I}$. 

\subsection{End-to-end fine-tuning}
Even though the training in each of the aforementioned three stages is performed separately, all the components are differentiable with respect to their parameters. Hence, Stochastic Gradient Descent (SGD) can be used to fine-tune the pipeline. 

Not all of the components are required for the fine-tuning, for instance the discriminator of the Adversarial Autoencoder is redundant. From the network in Stage I, only the encoder is utilized for extracting the latent representations, then linear regression (learned matrix $\bm{A}$) can be thought of as a linear fully-connected layer. From network in Stage III, all the components are kept. The overall architecture for fine-tuning is depicted in Fig.~\ref{fig:linear_dynamics_network_structure}.

\subsection{Prediction}
The structure of our three-stage pipeline is simplified for performing predictions. The image $\bm{i}^{(t)}$ is encoded (only the encoder of the network in Stage I is required); the resulting representation $\bm{d}^{(t)}$ is multiplied by $\bm{A}$ to obtain $\hat{\bm{d}}^{(t + x)}$, which is fed into the conditional GAN to synthesize a new image $\hat{\bm{i}}^{(t + x)}$. This procedure is visually illustrated in Fig.~\ref{fig:linear_dynamics_network_structure}, while more formally:

\begin{figure*}[!h]
    \centering
    \includegraphics[width=1\linewidth]{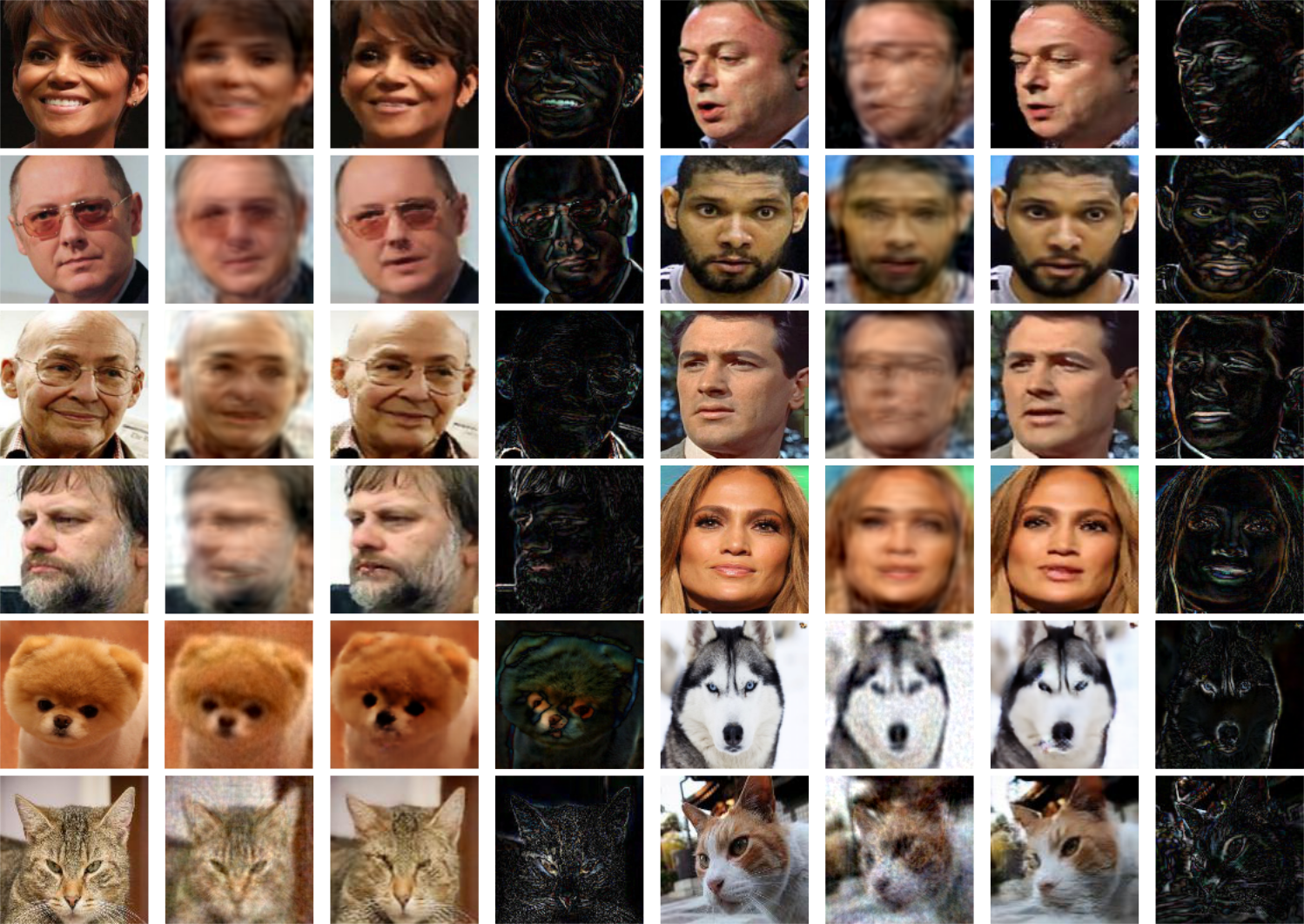}
    \caption{(Preferably viewed in color) Visual results of the synthesized images. There are four columns from the left to the right (split into left and right parts) which depict: (a) the original image, (b) the linear model (PCA + regression), (c) our proposed method, (d) the difference in intensities between the proposed method and the original image. The difference does not depict accurately the pose variation; the gif images in the supplementary material demonstrate the animated movement. Nevertheless, some noticeable changes are the following: a) in the left part in the second, and fifth images there is a considerable 3D rotation (pose variation), b) in the first, third and sixth in the left split there are several deformations (eyes closing, mouth opening etc.), c) in the second image on the right part, the person has moved towards the camera.}
    \label{fig:linear_dynamics_collage}
\end{figure*}

\begin{equation}
    \hat{\bm{i}}^{(t + x)} = f_d^{III}(f_e^{III}(\bm{A}\cdot[f_e^{I}(\bm{i}^{(t)}; 1)], \bm{i}^{(t)})))
\end{equation}

\subsection{Network architectures}
Our method includes two networks, i.e. an Adversarial Autoencoder for Stage I and a conditional GAN for Stage III. The encoder/decoder of both networks share the same architecture, i.e. 8 convolutional layers followed by batch normalization~\cite{ioffe2015batch} and LeakyRELU~\cite{maas2013rectifier}. The discriminator consists of 5 layers in both cases, while the dimensionality of the latent space is 1024 for all cases. Please refer to the table in the supplementary material for further details about the layers.

\section{Experiments}
\label{sec:linear_dynamics_experiment}

In this Section we provide the details of the training procedure along with the dedicated qualitative and quantitative results for all three objects, i.e. human faces, cats' faces and dogs' faces. Our objective is to demonstrate that this augmentation leads to learning invariances, e.g. deformations, not covered by commonly used techniques. 

\subsection{Implementation details}
The pairs of images required by the second and third stages, were obtained by sequential frames of that object. Different sampling of $x$ was allowed per frame to increase the variation. To avoid the abrupt changes between pairs of frames, the structural similarity (SSIM) of a pair was required to lie in an interval, i.e. the frames with i) zero, ii) excessive movement were omitted. 

Each of the aforementioned stages was trained separately; after training all of them, we have performed fine-tuning in the combined model (all stages consist of convolutions). However, as is visually illustrated in figures in the supplementary material there are minor differences in the two models. The results of the fine-tuned model are marginally more photo-realistic, which consists fine-tuning optional.

\subsection{Datasets}
\label{ssec:linear_dynamics_dataset}
A brief description of the databases utilized for training is provided below: 

\textbf{Human faces}: The recent dataset of MS Celeb~\cite{guo2016ms} was employed for Stage I (Sec.~\ref{ssec:linear_dynamics_stageI}). MS Celeb includes 8,5 million facial images of 100 thousand celebrities consisting it one of the largest public datasets for static facial images. In our case, the grayscale images were excluded, while from the remaining images a subset of 2 million random images was sampled. For the following two stages that require pairs of images the dataset of 300 Videos in-the-wild (300VW)~\cite{shen2015first} was employed. This dataset includes 114 videos with approximately 1 minute duration each. The total amount of frames sampled for Stage II  (Sec.~\ref{ssec:linear_dynamics_stageII}) is 13 thousand frames; 10 thousand frames are sampled for validation, while the rest are used for training the network in Stage III (Sec.~\ref{ssec:linear_dynamics_stageIII}).  

\textbf{Cat faces}: The pet dataset of \cite{parkhi2012cats} was employed for learning representations of cats' faces. The dataset includes 37 different breeds of cats and dogs (12 for cats) with approximately 200 images each\footnote{Each image is annotated with a head bounding box.}. In addition to those, we collected 1000 additional images, for a total of 2000 images. For the subsequent stages of our pipeline, pairs of images were required, hence we have collected 20 videos with an average duration of 200 frames. The head was detected with the DPM detector of \cite{felzenszwalb2010object} in the first frame and the rest tracked with the robust MDNET tracker of \cite{nam2016mdnet}. Since the images of cats are limited, the prior weights learned for the (human) facial experiment were employed (effectively the pre-trained model includes a prior which we adapt for cats).

\textbf{Dog faces}: The Stanford dog dataset~\cite{khosla2011novel} includes 20 thousand images of dogs from 120 breeds. The annotations are in the body-level, hence the DPM detector was utilized to detect a bounding box of the head. The detected images, i.e. 8 thousand, consisted the input for Stage I of our pipeline. Similarly to the procedure for cats, 30 videos (with average duration of 200 frames) were collected and tracked for Stages II and III.

\subsection{Variance in the latent space}
\label{ssec:linear_dynamics_variance_in_dynamics}
A quantitative self-evaluation experiment was to measure the variance of latent representations per video. The latent representations of sequential frames should be highly correlated; hence the variance in a video containing the same object should be low. 

A PCA was learned per video and the cumulative eigenvalue ratio was computed. We repeated the same procedure for all the videos (per object) and then averaged the results. The resulting plots with the average cumulative ratio are visualized in Fig.~\ref{fig:linear_dynamics_pca_variance}. In the videos of the cats and the dogs, we observe that the first 30 components express 90\% of the variance. In the facial videos that are longer (over 1500 frames) the variance is greater, however the first 50 components explain over 90\% of the variance.

\subsection{Qualitative assessment}
Considering the sub-space defined by PCA as the latent space and learning a linear regression there is the linear counterpart of our proposed method. To demonstrate the complexity of the task, we have learned a PCA per object\footnote{To provide a fair comparison PCA received the same input as our method (i.e. there was no effort to provide further (geometric) details about the image, the pixel values are the only input).}; the representations of each pair were extracted, linear regression was performed and then the regressed representations were used to create the new sample. 

In Fig.~\ref{fig:linear_dynamics_collage}, we have visualized some results for all three cases (human, cats' and dogs' faces). In all cases the images were not seen during the training with the cats' and dogs' images being downloaded from the web (all were recently uploaded), while the faces are from WIKI-DB dataset~\cite{rothe2016deep}. The visualizations verify our claims that a linear transformation in the latent space, can produce a realistic non-linear transformation in the image domain. In all of the facial images there is a deformation of the mouth, while in the majority of them there is a 3D movement. On the contrary, on the dogs' and the cats' faces, the major source of deformation seems to be the 3D rotation. An additional remark is that the linear model, i.e. regressing the components of PCA, does not result in realistic new images, which can be attributed to the linear assumptions of PCA. 

Aside of the visual assessment of the synthesized images, we have considered whether the new synthesized image is realistic enough to be considered as input itself to the pipelne. Hence, we have run an iterative procedure of applying our method, i.e. the outcome of iteration $k$ becomes the input to iteration $k + 1$. Such an iterative procedure essentially creates a collection of different images (constrained to include the same object of interest but with slightly different latent representations). Two such collections are depicted in Fig.~\ref{fig:linear_dynamics_sequential_executions}, where the person in the first row performs a 3D movement, while in the second different deformations of the mouth are observed. The image on the left is animated, hence if opened with Adobe Acrobat reader the transitions will be auto-played. We strongly encourage the reviewers to view the animated images and check the supplementary animations.

\subsection{Age estimation with augmented data}
To ensure that a) our method did not reproduce the input to the output, b) the images are close enough (small change in the representations) we have validated our method by performing age estimation with the augmented data. 

We utilized as a testbed the AgeDB dataset of \cite{moschoglou2017agedb}, which includes 16 thousand manually selected images. As the authors of \cite{moschoglou2017agedb} report, the annotations of AgeDB are accurate to the year, unlike the semi-automatic IMDB-WIKI dataset of \cite{rothe2016deep}. For the aforementioned reasons, we selected AgeDB to perform age estimation with i) the original data, ii) the original plus the new synthesized samples. The first 80\% of the images was used as training set and the rest as test-set. We augmented only the training set images with our method by generating one new image for every original one. We discarded the examples that have a structural similarity (SSIM)~\cite{wang2004image} of less than 0.4 with the original image; this resulted in synthesizing 6 thousand new frames (approximately 50\% augmentation). 

We trained a Resnet-50~\cite{he2015deep}  with i) the original training images, ii) the augmented images and report here the Mean Absolute Error (MAE). The pre-trained DEX~\cite{rothe2016deep} resulted in a MAE of 12.8 years in our test subset~\cite{moschoglou2017agedb}, the Resnet with the original data in MAE of 11.4 years, while with the augmented data resulted in a MAE of 10.3 years, which is a 9.5\% relative decrease in the MAE. That dictates that our proposed method can generate new samples that are not trivially replicated by affine transformations. 

\section{Conclusion}
\label{sec:linear_dynamics_conclusion}

In this work, we have introduced a method that finds a low-dimensional (approximately) linear space. We have introduced a three-stage approach that learns the transformations from the hihgly non-linear image space to the latent space along with the inverse transformation. This approach enables us to make linear changes in the space of representations and these result in non-linear changes in the image space. The first transformation was approximated by an Advervarsial Autoencoder, while a conditional GAN was employed for learning the inverse transformation and acquiring the synthesized image. The middle step consists of a simple linear regression to transform the representations. We have visually illustrated that i) the representations of a video form a discrete cluster (T-SNE in Fig.~\ref{fig:linear_dynamics_tsne})  ii) the representations of a single video are highly correlated (average cumulative eigenvalue ratio for all videos).

{\small
\bibliographystyle{ieee}
\bibliography{egbib}
}

\end{document}